\pdfoutput=1
\documentclass[11pt]{article}
\usepackage[final]{acl}
\usepackage{times}
\usepackage{latexsym}
\usepackage[T1]{fontenc}
\usepackage[utf8]{inputenc}
\usepackage{microtype}
\usepackage{inconsolata}
\usepackage{graphicx}
\usepackage{afterpage}
\usepackage{float}
\usepackage{multirow} 
\usepackage{booktabs}
\usepackage{colortbl}
\usepackage{xcolor}
\usepackage{amsmath}
\usepackage{amsfonts}  
\usepackage{amssymb}

\definecolor{lightgray}{rgb}{0.95, 0.95, 0.95}

\title{SongSage: A Large Musical Language Model with Lyric Generative Pre-training}

\author{
 \textbf{Jiani Guo\textsuperscript{1}\footnotemark[1]},
 \textbf{Jiajia Li\textsuperscript{2}\footnotemark[1]},
 \textbf{Jie Wu\textsuperscript{3}},
 \textbf{Zuchao Li\textsuperscript{4}\footnotemark[2]},
 \textbf{Yujiu Yang\textsuperscript{3}},
 \textbf{Ping Wang\textsuperscript{2}}
\\ 
 \textsuperscript{1}School of Computer Science, Wuhan University \\
 \textsuperscript{2}School of Information Management, Wuhan University \\
 \textsuperscript{3}Tsinghua University;
 \textsuperscript{4}School of Artificial Intelligence, Wuhan University \\
 \texttt{\{guojiani, cantata, zcli-charlie\}@whu.edu.cn} \\
\\
}

\begin{document}
\maketitle
\begin{abstract}
Large language models have achieved significant success in various domains, yet their understanding of lyric-centric knowledge has not been fully explored. 
In this work, we first introduce \textbf{PlaylistSense}, a dataset to evaluate the playlist understanding capability of language models.
PlaylistSense encompasses ten types of user queries derived from common real-world perspectives, challenging LLMs to accurately grasp playlist features and address diverse user intents.
Comprehensive evaluations indicate that current general-purpose LLMs still have potential for improvement in playlist understanding.
Inspired by this, we introduce \textbf{SongSage}, a large musical language model equipped with diverse lyric-centric intelligence through lyric generative pretraining.
SongSage undergoes continual pretraining on \textbf{LyricBank}, a carefully curated corpus of 5.48 billion tokens focused on lyrical content, followed by fine-tuning with \textbf{LyricBank-SFT}, a meticulously crafted instruction set comprising 775k samples across nine core lyric-centric tasks.
Experimental results demonstrate that SongSage exhibits a strong understanding of lyric-centric knowledge, excels in rewriting user queries for zero-shot playlist recommendations, generates and continues lyrics effectively, and performs proficiently across seven additional capabilities.
Beyond its lyric-centric expertise, SongSage also retains general knowledge comprehension and achieves a competitive MMLU score.
We will keep the datasets inaccessible due to copyright restrictions and release the SongSage and training script to ensure reproducibility and support music AI research and applications, the datasets release plan details are provided in the appendix.

\end{abstract}

\renewcommand{\thefootnote}{\fnsymbol{footnote}}
\footnotetext[1]{Equal contribution.}
\footnotetext[2]{Corresponding Author. }

\section{Introduction}
Artificial Intelligence (AI), especially Large Language Models (LLMs) have achieved significant success in various natural language processing tasks, including mathematical reasoning~\cite{gou2024critic,guo-etal-2025-tom,}, instruction following~\cite{DBLP:journals/corr/abs-2311-07911,10.1007/978-981-95-4088-4_15}, code generation~\cite{DBLP:conf/iclr/LuoX0SGHT0LJ24,wu-etal-2025-teaching,wu2025iterpreffocalpreferencelearning}, and multilingual processing~\cite{DBLP:conf/acl/UstunAYKDOBSOKV24,10.1145/3689090.3689389}.
Building on these successes, researchers have been exploring the potential of LLMs in understanding specialized domains beyond natural language, ranging from healthcare~\cite{singhal2023large} and chemistry~\cite{DBLP:journals/natmi/BranCSBWS24} to biology~\cite{madani2023large} and other professional fields.

Music, as a rich and multifaceted art form, spans multiple modalities: linguistic (lyrics), visual (musical scores), and auditory (vocals, melodies, accompaniments, and full compositions). 
Given this diversity, LLMs have been increasingly being explored in various forms of music-related research, including music generation~\cite{DBLP:conf/acl/HongHCWLYZZ24,DBLP:conf/naacl/MelechovskyGGMH24,hong-etal-2024-text,DBLP:conf/naacl/MelechovskyGGMH24}, musical knowledge understanding~\cite{yuan-etal-2024-chatmusician,deng-etal-2024-musilingo,DBLP:journals/corr/abs-2407-21531,Wu2024FUTGATF,weck2024muchomusicevaluatingmusicunderstanding,sakshi2024mmaumassivemultitaskaudio}, and musical score completion~\cite{li-etal-2024-music}.
Given that language has a higher information density compared to other modalities, LLMs are often used as the foundation for developing music AI models.

To evaluate whether general-purpose LLMs suffice for music AI, we introduce PlaylistSense, a 5k-playlist dataset designed to assess LLMs' understanding of playlists and lyrics. It simulates user queries based on ten real-life scenarios, testing LLMs' ability to capture playlist characteristics and generate precise descriptions that allow queries to retrieve the intended playlists.
Extensive evaluations indicate that LLMs exhibit limitations in playlist comprehension and lyrics understanding, along with various inherent biases, these findings underscore the critical need for specialized large musical language models (LMLMs) to advance the field of music AI.

Compared to general-purpose LLMs, LMLMs are typically further pretrained on music-specific textual content, predominantly comprising song lyrics. While song lyrics are extensively available, other forms of music-related texts are relatively scarce.
Inspired by this, we introduce SongSage, a large musical language model with lyric generative pretraining on a vast archive of song lyrics sourced from the internet.
SongSage is built upon two training datasets: \textit{LyricBank} and \textit{LyricBank-SFT}.
\textit{LyricBank}, our meticulously curated continual pretraining corpus, consists of 5.48 billion tokens, designed to enrich the model's understanding of lyrical content. 
Building on this foundation, \textit{LyricBank-SFT}, an instruction-tuning dataset comprising 775k samples, introduces nine specialized lyric-related capabilities, further refining SongSage's expertise in the music domain.
Following continual pretraining and instruction tuning, SongSage showcases exceptional music comprehension while also exhibiting fair general knowledge understanding. 
We believe that SongSage, an open-source foundation model with music knowledge, will facilitate further advancement in the field of large musical language models. Specifically, our contributions are:

(1) We propose the \textit{PlaylistSense} dataset focused on evaluating the playlist understanding capability of language models from ten user-query perspectives.

(2) We meticulously curate a continual pretraining corpus \textit{LyricBank} along with an instruction-tuning dataset \textit{LyricBank-SFT} focused on song-related capabilities, on which we introduce SongSage, 
a large musical language model equipped with versatile lyrics-centric intelligence through lyric generative pretraining. Experimental results show that SongSage significantly improves in understanding lyric-centric tasks, while maintaining a fair general knowledge understanding.

(3) We analyze the linguistic features of lyrics pretraining in LLMs to investigate the differences it introduces compared to general LLMs.
\section{Related work}

\subsection{Music Knowledge Understanding}
Recent work has advanced LLMs' understanding of music knowledge and theory. The ZIQI-Eval benchmark~\cite{li-etal-2024-music}, with over 14,000 curated questions, tests diverse musical knowledge, showing that only GPT-4~\cite{openai2024gpt4technicalreport} achieves moderate performance. ChatMusician~\cite{yuan-etal-2024-chatmusician} enhances LLMs with musical abilities through training on ABC notation, treating music as a second language to enable comprehension of musical concepts while retaining core language capabilities.

\subsection{Music Representation and Generation}
Recent approaches to music understanding and generation can be broadly classified into audio-based and symbolic-based methods.

Audio-based methods process music as continuous signals or quantized audio tokens. Models like AudioGen~\cite{DBLP:conf/iclr/KreukSPSDCPTA23} and MusicGen~\cite{DBLP:conf/nips/CopetKGRKSAD23} use autoregressive transformers to generate discrete audio tokens, while M²UGen~\cite{DBLP:journals/corr/abs-2311-11255} extends these capabilities to multi-modal generation, integrating audio, text, and visual elements to guide music creation.

Symbolic-based approaches use various text-based representations to encode musical information, each suited to different aspects of understanding and generation. For instance, ABC Notation, used by ChatMusician~\cite{yuan-etal-2024-chatmusician}, encodes pitch, duration, and meter in a compact format, like "F4" for the F note in the fourth octave. MusicAgent~\cite{yu-etal-2023-musicagent} adopts a MIDI-like text format, converting MIDI data into sequences with pitch, velocity, and timing. Score-based notation, seen in ZIQI-Eval and MusicAgent, translates traditional music scores into text, preserving structural elements. Similarly, SongComposer~\cite{DBLP:journals/corr/abs-2402-17645} combines note attributes with lyrics in structured tuples, aligning music and text. These representations allow symbolic methods to capture both structural and expressive aspects of music.

\begin{figure*}[!t]
    \centering \includegraphics[width=0.98\linewidth]{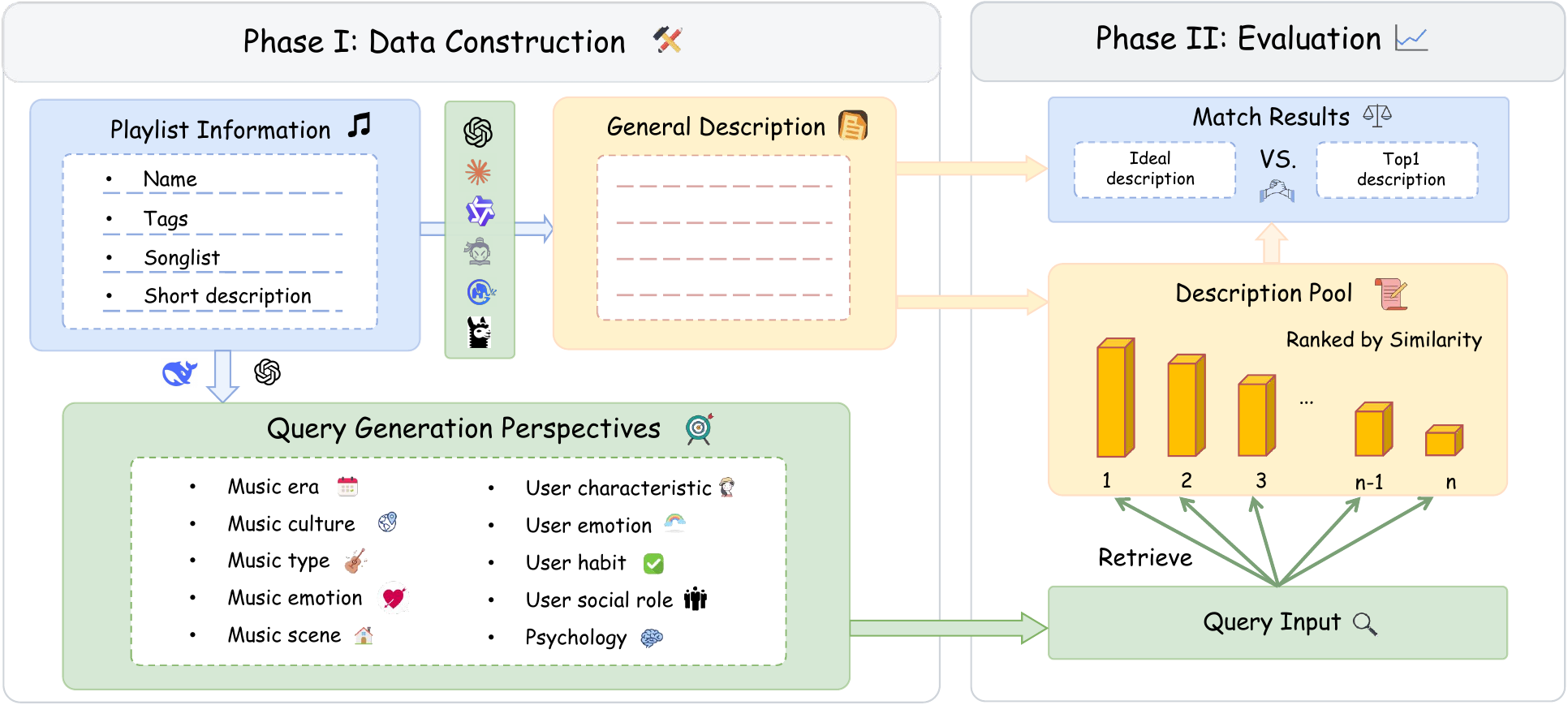}
    \caption{Overview of the PlaylistSense evaluation. In the data construction process, Deepseek generates 10 queries, covering both Music-Based and User-Based perspectives. During evaluation, the evaluator generate descriptions for each playlist, forming a pool. Each query retrieves descriptions from the pool via an embedding model, with success determined by whether the retrieved description matches the query's source playlist.}
    \label{benchmark}
\end{figure*}

\section{PlaylistSense Evaluation}

We introduce the PlaylistSense dataset to evaluate the capacity of general-purpose LLMs to comprehend playlists. The dataset includes 5,000 playlists from \textit{NetEase Music}\footnote{NetEase Music is a popular online music streaming platform with diverse playlists. See \url{https://music.163.com/}. Data was collected for academic use only.}, with each playlist accompanied by 10 queries generated by providing playlist details to advanced LLM: Deepseek-v2.5~\cite{deepseekai2024deepseekv2strongeconomicalefficient}. These queries encompass both music-based and user-based angles, representing typical real-world user inquiries, as illustrated in Figure~\ref{fig:query_examples}. Further details regarding the dataset construction are provided in the appendix.

\subsection{Evaluation}
During the evaluation phase, we evaluate several advanced open-source and closed-source LLMs using the following approach in the PlaylistSense. For the $i$-th playlist, we input its details (title, tags, and the first 20 song lyrics) into the evaluated LLM in a zero-shot setting to generate a general description $\text{des}_i$, with content exceeding the model's window size truncated.
This process can be formalized as:
\begin{equation*}
    \text{des}_i = \text{Evaluator} (\text{title}_i \, || \, \text{tags}_i \, || \, \text{song lyrics}_i)
\end{equation*}

These descriptions collectively form the description pool \( D \), which reflects the understanding of each playlist by the evaluator:
\begin{equation*}
    D = \{\text{des}_1, \text{des}_2, \ldots, \text{des}_N\}
\end{equation*}

We adopt a retrieval-based evaluation paradigm, using Hit Rate@$k$ (HR@$k$) as the metric.
The evaluation process works as follows: For each $\text{query}_i$, we retrieve the top $k$ most similar descriptions from the description pool $D$. 
HR@$k$ then calculates the proportion of cases where the correct description (corresponding to the same playlist as the query) appears within these top $k$ retrieved descriptions.
HR@$k$ can be formally defined as:
\begin{align*}
    \text{HR}@k &= \frac{1}{N} \sum_{i=1}^N \mathbb{I}\left(\text{des}_i \in \text{Top}_k^D(\text{query}_i)\right),
\end{align*}
\begin{align*}
    \text{Top}_k^D(\text{query}_i) &= \operatorname{Top}_k\left(\text{sim}\left(\mathbb{E}(\text{query}_i), \mathbb{E} (D)\right)\right),
\end{align*}

where $\text{Top}_k^D(\text{query})$ returns the top-$k$ most similar descriptions from the description pool $D$ based on cosine similarity between $\text{query}_i$ and all descriptions in $D$.
$\mathbb{I}(\cdot)$ is the indicator function, which equals 1 if the condition inside holds, and 0 otherwise.
$\mathbb{E}$ denotes an embedding model, bge-small-zh-v1.5~\cite{xiao2024cpackpackedresourcesgeneral} in practice.

We begin by evaluating GPT-4o, Claude 3.5 Sonnet~\cite{anthropic2024claude}, DeepSeek-V3, DeepSeek-R1~\cite{guo2025deepseek} and several open-source LLMs: Llama-3-8B-Instruct~\cite{dubey2024llama3}, Qwen2.5-7B-Instruct~\cite{yang2024qwen2}, GLM-4-9B-Chat~\cite{zeng2024chatglm}, and InternLM2.5-7B-Chat~\cite{cai2024internlm2}. The HR@1 and HR@5 results are presented in Table~\ref{benchmark_results}.

\begin{table*}[!t]
\centering
\renewcommand{\arraystretch}{1.1}  
\small
\resizebox{0.88\textwidth}{!}{%
\begin{tabular}{c|cc|cc|cc|cc|cc}
\toprule
\multicolumn{1}{c|}{} & \multicolumn{10}{c}{\textbf{Music-Based Perspective}} \\
\cmidrule(lr){2-11}
\textbf{Model} & \multicolumn{2}{c|}{Culture} & \multicolumn{2}{c|}{Emotion} & \multicolumn{2}{c|}{Era} & \multicolumn{2}{c|}{Scene} & \multicolumn{2}{c}{Type} \\
\cmidrule(lr){2-3} \cmidrule(lr){4-5} \cmidrule(lr){6-7} \cmidrule(lr){8-9} \cmidrule(lr){10-10}
& {HR@1} & {HR@5} & {HR@1} & {HR@5} & {HR@1} & {HR@5} & {HR@1} & {HR@5} & {HR@1} & {HR@5} \\
\midrule
\rowcolor{gray!10} Llama-3-8B-Instruct & 0.53 & 0.61 & 0.41 & 0.49 & 0.43 & 0.68 & 0.31 & 0.61 & 0.47 & 0.61 \\
InternLM2.5-7B-Chat & 14.78 & 28.94 & 8.85 & 19.53 & 14.20 & 19.61 & 9.43 & 26.73 & 12.63 & 0.02 \\
\rowcolor{gray!10} Qwen2.5-7B-Instruct & 14.49 & 28.55 & 9.47 & 20.11 & 14.16 & 16.27 & 8.18 & 17.46 & 12.26 & 25.99 \\
GLM4-9B-Chat & \underline{17.71} & \underline{34.97} & \underline{9.72} & \underline{21.73} & \textbf{16.85} & 19.94 & \underline{11.03} & \textbf{23.53} & \textbf{14.78} & 30.21 \\
\rowcolor{gray!10} Claude-3.5-Sonnet& 10.56 & 24.51 & 7.40 & 18.51 & 10.53 & 24.76 & 6.76 & 17.36 & 9.35 & 24.10 \\
GPT-4o & \textbf{18.94} & \textbf{35.99} & \textbf{18.90} & \textbf{24.12} & 12.1 & \textbf{35.03} & \textbf{17.52} & \underline{23.51} & 11.48 & \textbf{33.59} \\
\rowcolor{gray!10} DeepSeek-V3& 15.70 & 32.34 & 8.51& 19.80 & \underline{15.47} & \underline{31.17} & 8.96 & 19.70 & \underline{14.06} & \underline{30.42} \\
DeepSeek-R1 & 2.28 & 6.44 & 1.87 & 5.66 & 3.03 & 8.32 & 1.68 & 5.41 & 2.95 & 8.08 \\
\midrule[\heavyrulewidth]
\multicolumn{1}{c|}{} & \multicolumn{10}{c}{\textbf{User-Based Perspective}} \\
\cmidrule(lr){2-11}
\textbf{Model} & \multicolumn{2}{c|}{Characteristic} & \multicolumn{2}{c|}{Emotion} & \multicolumn{2}{c|}{Habit} & \multicolumn{2}{c|}{Social Role} & \multicolumn{2}{c}{Psychology} \\
\cmidrule(lr){2-3} \cmidrule(lr){4-5} \cmidrule(lr){6-7} \cmidrule(lr){8-9} \cmidrule(lr){10-10}
& {HR@1} & {HR@5} & {HR@1} & {HR@5} & {HR@1} & {HR@5} & {HR@1} & {HR@5} & {HR@1} & {HR@5} \\
\midrule
\rowcolor{gray!10} Llama-3-8B-Instruct & 0.41 & 0.55 & 0.14 & 0.47 & 0.18 & 0.31 & 0.33 & 0.43 & 0.08 & 0.18 \\
InternLM2.5-7B-Chat & 9.47 & 20.91 & 6.13 & 13.79 & 5.51 & 12.56 & 10.49 & 21.95 & 1.23 & 3.79 \\
\rowcolor{gray!10} Qwen2.5-7B-Instruct & 8.51 & 18.82 & 6.50 & 13.92 & 5.29 & 11.46 & 10.58 & 21.68 & 1.15 & 3.30 \\
GLM4-9B-Chat & \underline{11.85} & \underline{25.50} & \underline{6.58} & \underline{16.03} & 5.74 & \underline{13.18} & \underline{11.91} & \underline{25.29} & \textbf{1.50} & \textbf{4.39} \\
\rowcolor{gray!10} Claude-3.5-Sonnet & 8.20 & 20.68 & 4.57 & 14.02 & 4.12 & 10.31 & 7.77 & 19.23 & 1.05 & 3.79 \\
GPT-4o & \textbf{13.38} & \textbf{25.64} & \textbf{6.97} & \textbf{16.60} & \textbf{7.26} & \textbf{13.86} & \textbf{13.10} & \textbf{25.85} & \underline{1.35} & \underline{3.82} \\
\rowcolor{gray!10} DeepSeek-V3& 10.06 & 23.26 & 4.48 & 12.24 & \underline{5.82} & 12.95 & 9.69 & 22.44 & 0.88 & 2.40 \\
DeepSeek-R1 & 2.09 & 6.23 & 1.05 & 3.85 & 1.58 & 4.80 & 1.72 & 6.25 & 0.23 & 1.09 \\
\bottomrule
\end{tabular}
}
\caption{HR@1(\%) and HR@5(\%) of GPT-4o, Claude 3.5 Sonnet, DeepSeek-V3, DeepSeek-R1 and advanced open-source models across 10 perspectives on the PlaylistSense dataset in Chinese. Bold denotes the best performance, and underlined results indicate runner-ups.}
\label{benchmark_results}
\end{table*}

Based on our comprehensive evaluation of multiple LLMs across ten distinct perspectives, we present the following two key observations and insights.

\subsection{Observation \#1: LLMs Face Significant Limitations in Understanding Playlists.}
Current LLMs still exhibit insufficient understanding of playlists, both in open-source and closed-source ones.
Even advanced LLMs like GPT-4o and Claude-3.5-Sonnet do not exhibit a clear performance advantage. 

GPT-4o almost leads in both \textit{Music-Based} and \textit{User-Based} perspectives, but its performance is still limited, with only 18.94\% HR@1 in the \textit{Music Culture}, its strongest perspective. GLM4-9B-Chat performs best among open-source models, scoring 16.85\% (HR@1) in the \textit{Music Era}, the highest.

Llama-3-8B-Instruct performs poorly, highlighting difficulties in playlist understanding, with only marginal improvement in HR@5 compared to HR@1. Surprisingly, DeepSeek-R1, despite being considered as one of the state-of-the-art LLMs, ranks as the second-worst.

This underscores the significant limitations of current LLMs in understanding playlists, which remain far from ideal.
The challenge becomes even more pronounced on real-world playlist platforms, where the number of playlists vastly exceeds the 5,000 included in PlaylistSense.

\subsection{Observation \#2: There are Biases in Playlist Understanding.}

\textbf{I. Perspective Bias.}
The LLMs generally exhibit consistent trends across various perspectives, with superior performance in the \textit{Music-Based}  compared to the \textit{User-Based} perspective. For example, DeepSeek-V3 achieves the highest HR@1 score of 15.70\% in \textit{Music Culture}, yet performs poorly in \textit{Psychology} with only 0.88\%. Particularly in \textit{Psychology}, LLMs exhibit poor performance.

\textbf{II. Comprehension Depth Bias.}
Relatively speaking, LLMs perform better with surface-level features related to music, like \textit{Music Culture} and \textit{Music Era}, than with deeper user-related aspects such as \textit{Psychology} and \textit{User Emotion}.
According to the results, GLM4-9B-Chat scores the highest in \textit{Music Culture} (HR@1: 17.71\%), far surpassing its scores in \textit{Psychology} (HR@1: 1.50\%). 
This phenomenon may shed light on the limitations of most LLMs in understanding playlists through the lens of deep user psychology and emotions. We will explore this issue further in our future research.

\textbf{III. Cross-dimensional Stability Bias.}
Even within the same category of understanding, such as the Music-Based perspective, LLMs exhibit significant performance fluctuations across different dimensions. For example, GPT-4o achieves 18.94\% (HR@1) in \textit{Music Culture}, but drops to 11.48\% (HR@1) in \textit{Music Type}, highlighting the instability in LLMs' cross-dimensional performance, where strong performance in some dimensions may substantially drop in others, affecting overall stability.
\section{LMLMs: SongSage}

\subsection{Overview}

\begin{figure*}[!t]
    \centering \includegraphics[width=0.98\linewidth]{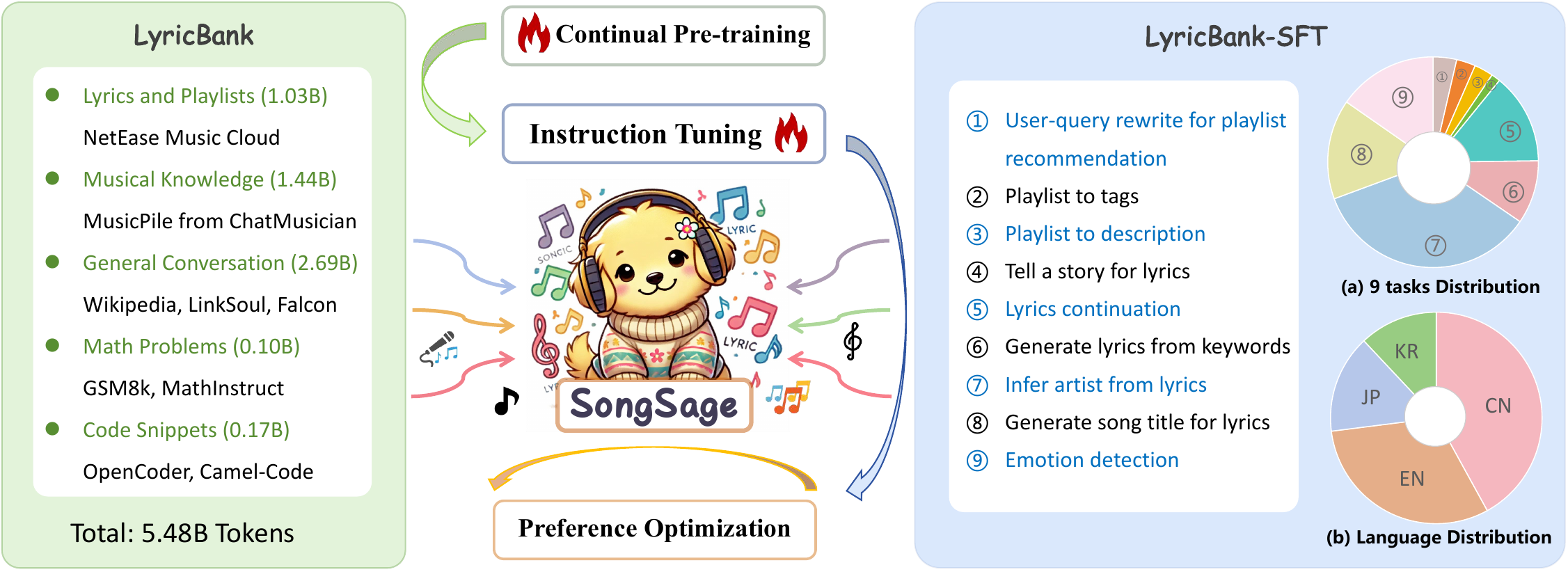}
    \caption{Overview of SongSage's three-stage training, detailing the progression from general lyrical knowledge on LyricBank to task-specific instruction tuning with LyricBank-SFT, followed by preference optimization, alongside a breakdown of knowledge domains and task distributions. The logo of SongSage is created using GPT-4.}
    \label{dataset}
\end{figure*}
Given that musical text fundamentally represents a form of linguistic description, we leverage  LLMs pretrained on general text as our foundation. Due to the current limitations in LLMs' understanding of playlists, we aim to bridge this gap and improve their proficiency through targeted post-training efforts.
As illustrated in Figure~\ref{dataset}, we comprise three sequential stages: continual pretraining, instruction tuning, and preference optimization, each backed by a carefully curated dataset.

For this purpose, we construct datasets containing millions of lyrical samples, including LyricBank, LyricBank-SFT, and a curated collection of preference data.
\textbf{LyricBank} aims to enhance the LLM's understanding of lyrical content with generative continual pretraining, while \textbf{LyricBank-SFT} activates accumulated domain-specific knowledge through nine specific tasks on lyrics and playlists. \textbf{Preference set} enables further optimization of model performance and allows for tailored customization to meet specific needs.
We detail the composition of the three datasets.

\subsection{LyricBank}

We introduce LyricBank, a pretraining dataset containing 
200K playlists (including playlist names, tags, songlist), and lyrical information from 2.5M songs on NetEase Music (including details about singers, albums, and primarily the lyrics).

To ensure data quality, we deduplicate the data and remove tracks with minimal lyrical content, such as purely instrumental compositions. To maintain diversity, the dataset spans four languages: English, Chinese, Korean, and Japanese, in an approximately 2:2:1:1 ratio. After filtering, the final collection comprises 1.8M song lyrics.

To prevent overfitting on lyrical data, we supplement LyricBank with a mix of general-purpose data to preserve the LLM's original general knowledge.
Figure~\ref{dataset} shows the proportion of different corpora,
including daily conversations, code snippets, math problems, and music knowledge data, with lyrical data maintaining a 1:4 ratio to the combined total of other data types.

\subsection{LyricBank-SFT}
To further leverage the lyrical knowledge accumulated in LyricBank, we develope LyricBank-SFT, an instruction-tuning dataset comprising nine key tasks centered around understanding and generating playlist and lyric content. 

These tasks are detailed in Figure~\ref{dataset} and are carefully designed to leverage the lyrical knowledge from LyricBank, covering both playlist-level and lyric-level understanding and generation.
They include practical applications such as playlist recommendations and creative tasks like generating stories from lyrics and composing new lyrics based on specified keywords and styles. To ensure data quality and provide better instructions to the model, we employ a multiround sampling and inspection method to verify the data.

LyricBank-SFT consists of 775k samples across nine task panels, with the data distribution for each capability illustrated in center of Figure~\ref{dataset}. 

To mitigate overfitting, we include MusicPile-sft\cite{yuan-etal-2024-chatmusician} to enhance musical knowledge, and Infinity-Instruct-0625~\cite{zhang2024inifinitymath} to improve skills in mathematics, coding, and instruction following.

\subsection{Preference Set}
We also construct a preference dataset for optimization.
As collecting real user preference data is challenging and ethically sensitive, we use responses from a stronger model (Deepseek for Chinese, GPT-4o for English) as preferred and from a weaker model (Yi-1.5-6B-Chat for Chinese, Llama3.1 for English) as rejected.
For each capability in nine SFT tasks, we generate 1k data points, which are used for Direct Preference Optimization (DPO)~\cite{DBLP:conf/nips/RafailovSMMEF23} after instruction tuning.
\section{Experiment}

\textbf{Training Configuration.}
We initialize SongSage-Base using a general foundation model, Qwen2.5-7B-Base~\cite{yang2024qwen2} weights, and adopt a continual pretraining followed by instruction-tuning pipeline, employing a maximum sequence length of 2048. During continual pretraining on LyricBank, the learning rate is set to 3e-5, with 0.8M tokens per batch.
Following ChatMusician, we adopt LoRA~\cite{DBLP:conf/iclr/HuSWALWWC22} for efficient instruction tuning, using a learning rate of 1e-4 and a batch size of 1024.
Training is conducted on 16 A100 GPUs, with one epoch for both continual pretraining and fine-tuning.
During the DPO stage, we use a batch size of 128 and train LoRA for 3 epochs. More detailed parameters and loss curves are provided in the Appendix.

\subsection{Evaluation of Playlist-Centric Tasks.}
Table~\ref{tab:sft-results} presents the performance of SongSage and GPT-4o on three playlist-centric tasks.
In the query rewrite task, we use SongSage to rewrite queries from PlaylistSense and retrieve results from a 300k Golden Descriptions library generated by Deepseek for each playlist. Compared to GPT-4o, the rewritten queries achieve a 28.4\% increase in average embedding similarity (Emb-Sim) with their corresponding golden descriptions. Additionally, SongSage improves Recall@1 and Recall@5 by 2.24\% and 3.47\%, respectively, marking a significant gains in this challenging recommendation scenarios where LLMs have access to only user queries.
Further improvements in the Playlist-to-Tags and Playlist-to-Description tasks on 10k samples show SongSage's ability to understand and capture playlist characteristics.

\begin{table*}[htbp]
\centering
\small
\begin{tabular}{l|@{\hspace{6pt}}ccc@{\hspace{6pt}}|@{\hspace{6pt}}cc@{\hspace{6pt}}|@{\hspace{6pt}}c@{\hspace{6pt}}}
\toprule
\multirow{2}{*}{Model} & \multicolumn{3}{c|}{User-query Rewrite} & \multicolumn{2}{c|}{Playlist to Tags} & \multicolumn{1}{c}{Playlist to Description} \\
\cmidrule(lr){2-4} \cmidrule(lr){5-6} \cmidrule(lr){7-7}
& R@1 $\uparrow$ & R@5 $\uparrow$ & Emb-Sim $\uparrow$ & Precision $\uparrow$ & F1-Score $\uparrow$ & Emb-Sim $\uparrow$ \\
\midrule
GPT-4o & 11.45  & 23.25  & 60.52 & 4.30 & 2.30 & 60.52 \\
SongSage (SFT) & \textbf{13.69} & \textbf{26.72}
  & \textbf{88.92}  & \textbf{45.8} & 34.8 & \textbf{89.51} \\
SongSage (DPO) & 13.15 & 24.20 & 83.64 & 40.6  & \textbf{37.9} & 86.84 \\
\bottomrule
\end{tabular}

\vspace{1em}

\begin{tabular}{l|cc|cc|c}
\toprule
Model & \multicolumn{2}{c|}{Infer Artist} & \multicolumn{2}{c|}{Generate Song Title} & Emotion Detection \\
\cmidrule(lr){2-3} \cmidrule(lr){4-5} \cmidrule(lr){6-6}
& Exact Match $\uparrow$ & Partial Match $\uparrow$ & F1-Score $\uparrow$ & Edit Distance $\downarrow$ & Accuracy $\uparrow$ \\
\midrule
GPT-4o & 9.65 & 1.80 & 16.3 & 6.17 & 31.65 \\
SongSage (SFT) & \textbf{21.55} & \textbf{4.70} & \textbf{56.37} & \textbf{3.41} & 32.90 \\
SongSage (DPO) & 18.45  & 4.20 & 55.17 & 3.57 & \textbf{35.70} \\
\bottomrule
\end{tabular}

\caption{
Results of SongSage on playlist-centric (top) and lyric-centric (bottom) tasks. Best scores in bold.}
\label{tab:sft-results}
\end{table*}

\subsection{Evaluation of Lyric-Centric Tasks.}
For lyric understanding, Table~\ref{tab:sft-results} presents the average performance of Infer Artist, Generate Song Title, and Emotion Detection tasks over 10k samples, where SongSage consistently outperforms GPT-4 in understanding lyrics. 

For lyric generation tasks, we first adopt the LLM-as-judge approach, with GPT-4o assessing the results of SongSage and Claude-3.5-Haiku across four dimensions: Fluency, Relevance, Creativity, and Emotional Expression. Detailed criteria and prompts are in the Appendix. Table~\ref{three-tasks} shows the average scores on 2k samples, with SongSage demonstrating better proficiency than Claude-3.5.

Further, we conduct a questionnaire focusing on two specific tasks: lyrics continuation and lyrics creation based on keywords. 
Each questionnaire contains 20 questions for both tasks. Participants are invited to assess which model produces the highest quality lyrics among SongSage, GPT-4o, Claude-3.5, and other models. A total of 110 questionnaires are collected, with participants including university students, middle-aged individuals, and over 30 professionals who major in music.

\begin{table}[!t]
\centering
\small
\begin{tabular}{l|cccc|c}
\toprule
\multicolumn{6}{c}{Keywords to Lyrics} \\
\midrule
Model & Creat & Fluen & Rele & Emo & \textbf{Avg.} \\
\midrule
Claude-3.5 & 7.91 & \textbf{8.83} & \textbf{9.30} & 8.42 & 8.61 \\
SongSage & \textbf{8.83} & 8.74 & 8.78 & \textbf{8.90} & \textbf{8.81} \\
\midrule
\multicolumn{6}{c}{Lyrics Continuation} \\
\midrule
Claude-3.5 & 5.15 & 5.91 & 5.54 & 5.76 & 5.59 \\
SongSage & \textbf{7.72} & \textbf{8.04} & \textbf{8.11} & \textbf{7.98} & \textbf{7.96} \\
\midrule
\multicolumn{6}{c}{Story Generation} \\
\midrule
Claude-3.5 & 5.24 & 7.44 & 5.59 & 5.23 & 5.92 \\
SongSage & \textbf{7.13} & \textbf{8.66} & \textbf{7.30} & \textbf{7.71} & \textbf{7.70} \\
\bottomrule
\end{tabular}
\caption{Performance across three generation tasks.}
\label{three-tasks}
\end{table}

\begin{table}[!t]
\centering
\small
\begin{tabular}{lcc}
\toprule
Model & Lyrics Continuation & Lyrics Creation \\
\midrule
GPT-4o & 20.08 & 23.50 \\
Claude-3.5 & 10.42 & 12.35 \\
GLM4-9B & 17.80 & 14.67 \\
SongSage & \textbf{42.12} & \textbf{33.38} \\
Qwen2.5-7B & 9.58 & 16.01 \\
Total & 100.0 & 100.0 \\
\bottomrule
\end{tabular}
\caption{Preference ratio of different models for lyrics continuation and creation tasks, based on 110 questionnaires, with 20 samples per task.}
\label{tab:questionnaires}
\end{table}

Table~\ref{tab:questionnaires} presents the human evaluation results. SongSage achieves significant preference gains for both tasks, with a 22.0\% higher preference ratio in lyrics continuation and an 9.9\% preference advantage in lyrics creation  over GPT-4o.
This preference demonstrates that, through lyric generative pre-training and targeted fine-tuning, SongSage is capable of functioning as a specialized large musical language model, effectively understanding real-world lyric preferences and generating high-quality, meaningful lyrics.

\subsection{Understanding of Playlists.}
\begin{figure}[!ht]
    \centering
    \includegraphics[width=1.0\linewidth]{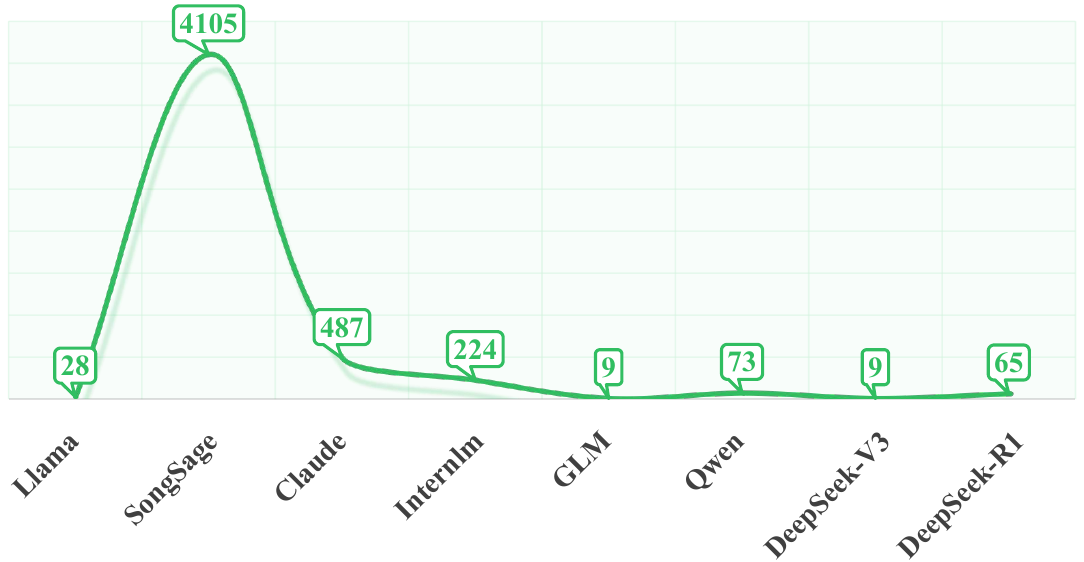}
    \caption{Performance of LLMs in generating playlist descriptions, with GPT-4o as the judge model. The numbers represent the number of times the model was selected.}
    \label{fig:description_judge}
\end{figure}
We use 5,000 playlist entries from PlaylistSense as input for the evaluated LLMs (same as Table~\ref{benchmark_results}, except GPT-4o) to generate descriptions. Adopting the LLM-as-judge approach, GPT-4o selects the best description across four dimensions: Accuracy, Emotional Conveyance, Information Coverage, and Scenario Suitability.
 Figure~\ref{fig:description_judge} shows the performance of LLMs, with SongSage leading at 4,105 selections, followed by Claude-3.5-Sonnet (487) and InternLM2.5-7B-Chat (224), highlighting SongSage's significant advantage in understanding playlist information.

\subsection{Understanding of Musical Knowledge.}
\begin{figure}[!ht]
    \centering
    \includegraphics[width=0.9\linewidth]{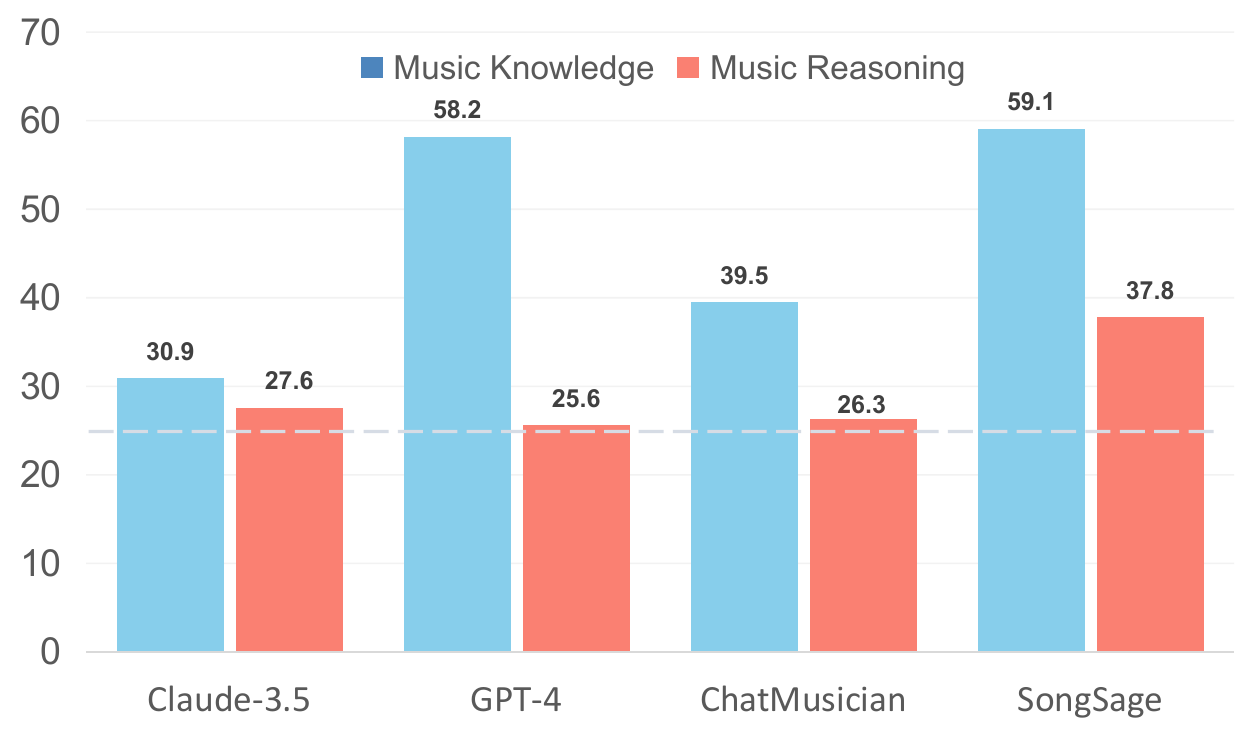}
    \caption{Accuracy(\%) on MusicTheoryBench. The horizontal dashed line represents the random guess score.}
    \label{fig:Music_theroybench}
\end{figure}

We report the performance of SongSage on MusicTheoryBench~\cite{yuan-etal-2024-chatmusician} and ZIQI-eval~\cite{li-etal-2024-music}.
Figure~\ref{fig:Music_theroybench} shows that SongSage achieves the best performance in the Music Knowledge task (59.1\%), outperforming ChatMusician by 19.6\%. Notably, SongSage exhibits significant performance gains in the Music Reasoning task (37.8\%), while other models remain near random-guess performance (25\%).
\begin{table}[!h]
\centering
\small
\begin{tabular}{lcc}
\hline
\textbf{Models}      & \textbf{Accuracy (\%)} & \textbf{F1-score (\%)} \\ \hline
ChatMusician         &  24.6                       & 23.9                \\
GPT-3.5-Turbo             & 50.2                      & 52.9           \\
GPT-4              &  \underline{62.9}                    & \underline{63.0}          \\
GPT-4o              & \textbf{76.1}                      & \textbf{76.1}           \\
SongSage             &  62.4                          & 62.1               \\ \hline
\end{tabular}
\caption{Performance comparision on ZIQI-Eval.}
\label{tab:ziqi_eval}
\end{table}
Table~\ref{tab:ziqi_eval} compares the performance of music comprehension test on ZIQI-Eval.
Although GPT-4o
performs relatively well, SongSage outperforms the base model, ChatMusician, and GPT-3.5-Turbo, closely follows GPT-4, with an accuracy of 62.4\% and an F1-score of 62.1\% in terms of the ZiQi Eval score.
Compared to GPT-3.5-Turbo and ChatMusician, SongSage outperforms them by 12.2\% and 37.8\% in accuracy, respectively, demonstrating its strong ability to comprehend and process musical knowledge.

\subsection{General Knowledge Understanding.}
\begin{table}[!ht]
\centering
\small
\begin{tabular}{lc}
\hline
\textbf{Model}           & \textbf{MMLU Score(\%)} \\ \hline
ChatMusician-Base        & 48.5                \\
ChatMusician             & 46.8                \\
Qwen-2.5-7B-Base         & 7\textbf{4.2}        \\ 
SongSage-Base            & 64.5                    \\
SongSage-Instruct        & 63.6                \\ 
SongSage-DPO             & 63.0                    \\ \hline
\end{tabular}
\caption{Performance on MMLU.}
\label{tab:mmlu_results}
\end{table}

Table~\ref{tab:mmlu_results} shows the performance on massive multitask language understanding (MMLU)~\cite{DBLP:conf/iclr/HendrycksBBZMSS21}.
SongSage-Base achieves a MMLU score of 64.5\%, lower than Qwen2.5-7B-base on MMLU bench because
LyricBank biases the LLM toward music-related knowledge, causing some loss of general
knowledge, and outperforming ChatMusician-Base by 16.0\%. 
This improvement benefits from the inclusion of general knowledge considerations in LyricBank and LyricBank-SFT.
ChatMusician exhibits weaker performance, likely due to two factors: its emphasis on learning the symbolic language of ABC Notation, which may limit general knowledge, and its foundation on Llama2-7B-Base, an earlier model with constrained capabilities.

\subsection{Linguistic Analysis of LLMs with Lyric Pretraining}
It would be interesting to investigate the linguistic differences introduced by lyric continual pretraining in LLMs. To this end, we utilize the spaCy syntactic parser~\cite{neumann2019scispacy} to analyze 1000 lyrics, which are generated by Qwen2.5-Base and SongSage-Base respectively, and examine the syntactic features of POS and Dependency distributions.

\begin{figure}[!ht]
    \centering
    \includegraphics[width=1.0\linewidth]{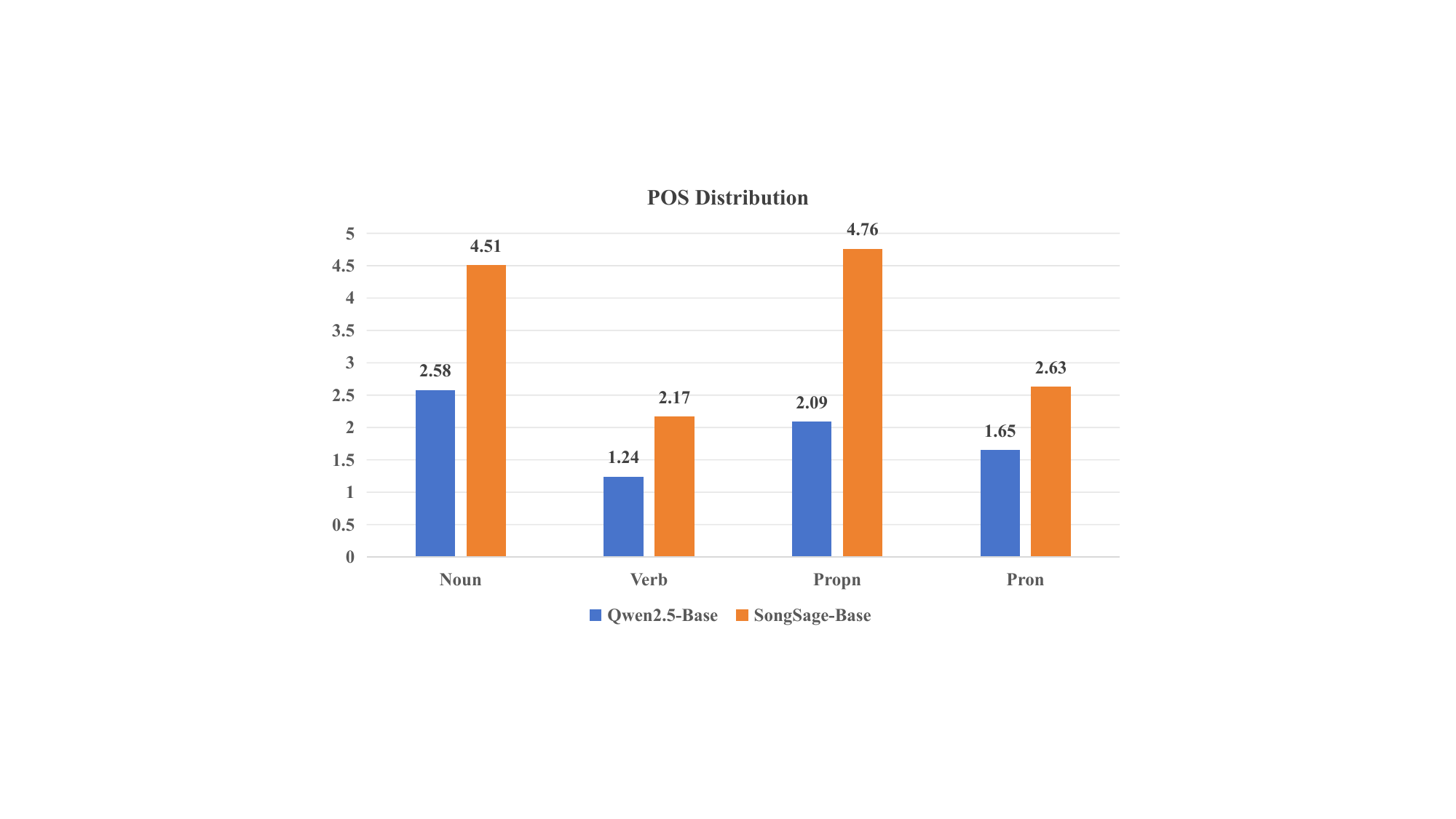}
    \caption{POS frequency distribution in Qwen2.5-Base and SongSage-Base. The numbers represent the average occurrences of each POS per sentence.}
    \label{fig:pos distribution}
\end{figure}

\textbf{I. Part-of-Speech (POS) Distribution Analysis.}
Figure~\ref{fig:pos distribution} shows that SongSage-Base outperforms Qwen2.5-Base in nouns (4.76\% vs. 2.09\%) and proper nouns (4.76\% vs. 2.09\%), indicating lyric pretraining encourages more specific and concrete language. Additionally, SongSage-Base has higher verb (2.17\% vs. 1.24\%) and pronoun (2.63\% vs. 1.65\%) frequencies, suggesting a greater emphasis on emotional expression and personalization, with pronouns enhancing lyric relatability and interactivity.

\begin{figure}[!ht]
    \centering
    \includegraphics[width=1.0\linewidth]{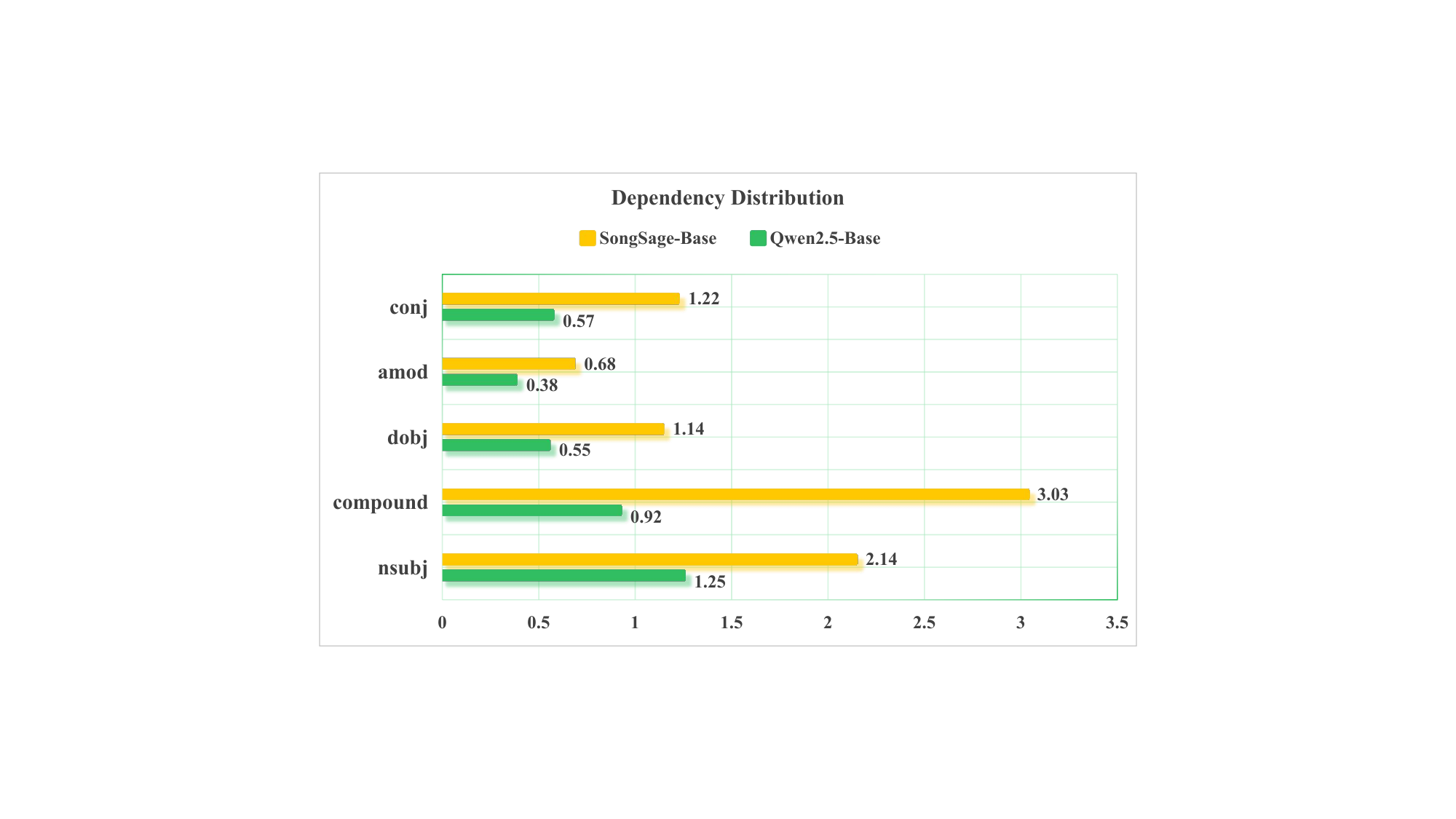}
    \caption{Dependency frequency distribution in Qwen2.5-Base and SongSage-Base. The numbers represent the average occurrences of each dependency per sentence.}
    \label{fig:dependency distribution}
\end{figure}

\textbf{II. Dependency Structure Analysis.}

Figure~\ref{fig:dependency distribution} shows that SongSage-Base outperforms Qwen2.5-Base in conjunctions (1.22\% vs. 0.57\%), adjectival modifiers (0.68\% vs. 0.38\%), and direct objects (1.14\% vs. 0.55\%), indicating a greater use of coordinating structures and emotional modifiers. Additionally, SongSage-Base's higher frequency of compound words (3.03\% vs. 0.92\%) suggests increased linguistic creativity and flexibility, typical of lyric writing.

\section{Conclusion}

Our research highlights the musical knowledge understanding capabilities of large language models. 
To evaluate whether general-purpose LLMs are adequate for music AI, we introduce PlaylistSense, a dataset designed to assess LLMs' understanding of playlists and lyrics. 
Our extensive evaluations reveal notable gaps in LLMs' ability to grasp these domains effectively. 
In response, we develop a large musical language foundation model with lyric generative pretraining. 
We also analyze the linguistic features of lyrics pretraining in LLMs, investigating the differences it introduces compared to general LLMs. 
As a result, we present LyricBank, a large-scale lyrical pretraining dataset designed to enhance comprehension of lyrical content, and LyricBank-SFT, an instruction-tuning dataset covering nine core lyric-based capabilities. 
Consequently, we introduce SongSage, a 7B-parameter LMLMs specialized in music-centered knowledge. Together with LyricBank and LyricBank-SFT, this suite of resources advances the state of music AI research.

\section*{Limitations}
In this work, we utilized all accessible lyric data for generative pretraining and optimized SongSage's capabilities within our resource constraints. While the current 7B-parameter model demonstrates superior performance compared to larger general-purpose models, we anticipate that scaling to larger model sizes will yield even more pronounced advantages in music-specific tasks.  Currently, our model supports English, Chinese, Japanese, and Korean; expanding to additional languages would further advance global music AI development. 
Looking ahead, we are actively developing a multimodal version of SongSage. While the current availability of music multimodal data is limited, we are continuously expanding our collection. Building upon SongSage's strong foundation in language understanding, we expect that incorporating additional multimodal music data will significantly enhance the model's comprehensive music understanding capabilities across various modalities.
\bibliography{ref}

\section*{Appendix}

\subsection*{Datasets Release Plan}

The metadata of our datasets—\textit{PlaylistSense}, \textit{LyricBank}, \textit{LyricBank-SFT}, and the \textit{preference dataset}—comprises information on playlists (such as titles, tags, and song lists) and songs (including song names, albums, and lyrics), which is sourced from \textit{NetEase Music}\footnote{NetEase Music is a popular online music streaming platform with diverse playlists. See \url{https://music.163.com/}. 
Data was collected for academic use only.}. We would like to emphasize that these datasets are solely for academic purposes and are not intended for any commercial use.

Due to data copyright restrictions, we regret to inform that we are currently unable to release the entirety of the datasets. Specifically, for PlaylistSense, we will make publicly available a portion of the data that has been artificially synthesized, which includes 10 different queries for each of the 5,000 Chinese playlists and 5,000 English playlists, representing various perspectives. The playlists themselves will not be made available. This subset of the data is provided in the attached file. All other datasets will remain unavailable.

Furthermore, we will release the model weights for SongSage, including SongSage-base, SongSage-Instruct, and SongSage-DPO, along with the corresponding training script.

\subsection*{PlaylistSense Dataset Construction.}

PlaylistSense includes 5,000 playlists in both Chinese and English from \textit{NetEase Music}.
Each playlist is accompanied by 10 queries designed to evaluate the model's understanding from distinct perspectives.
These queries are generated by providing playlist details, including the title, tags, and song list, to advanced LLMs: Deepseek-v2.5~\cite{deepseekai2024deepseekv2strongeconomicalefficient} for Chinese playlists and GPT-4o for English playlists. The LLMs are prompted separately across 10 perspectives to generate queries that emulate human inquiry styles.

\begin{figure}[!t]
    \centering
    \includegraphics[width=0.95\linewidth]{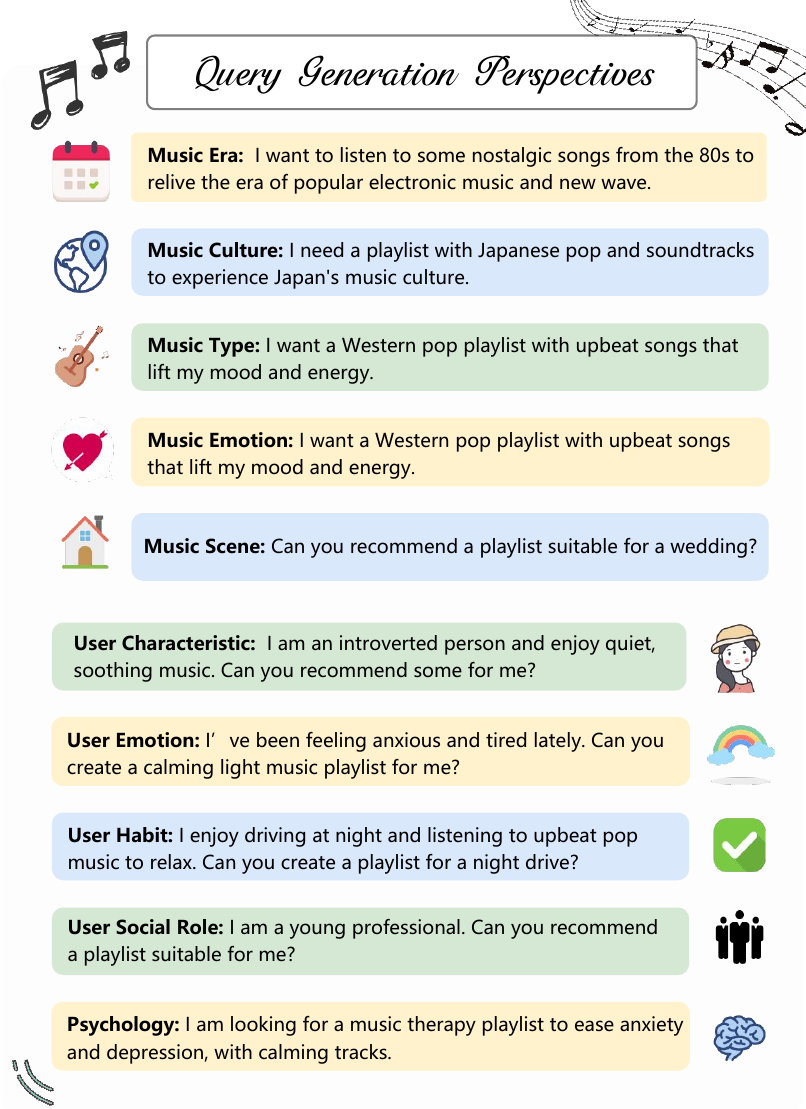}
    \caption{Examples of the ten types of queries.}
    \label{fig:query_examples}
\end{figure}

These 10 perspectives cover music-based and user-based angles, reflecting the types of inquiries users might make in real life.
As illustrated in Figure~\ref{fig:query_examples}, from the \textit{Music-Era} perspective, one might inquire, “I’d love to hear more music from the Beatles era, as well as songs from the 1960s cultural revolution.”
From the \textit{User-Psychology} perspective, a person suffering from depression might inquire, “Could you recommend a playlist that helps calm my nerves and reduce stress?”

To ensure the quality of these queries, we employ a few-shot approach, using human-evaluated examples which are provided by college students major in music to guide the model in generating more natural and human-like queries.

For the $i$-th playlist in the PlaylistSense benchmark, the $\text{query}_i^j$ from perspective $j$ is generated by feeding the playlist title, tags, and the first 20 song lyrics from the playlist into the LLM, Deepseek-v2.5 for Chinese or GPT-4o for English playlists, using a 5-shot approach. This can be expressed as:
\begin{align*}
    \text{query}_i^j = \text{LLM}(&\text{instruction}^j, \text{few-shot}^j; \\
    &\text{title}_i \, || \, \text{tags}_i \, || \, \text{song lyrics}_i)
\end{align*}

Here, the index $j$ corresponds to different perspectives (e.g., user emotion, music era, etc.).
The query set for perspective $j$ is defined as: 
\begin{align*}
    Q^j = \{\text{query}_1^j, \text{query}_2^j, \ldots, \text{query}_N^j\}, \\
    j \in \{1, 2, \ldots, 10\},
\end{align*}
where $N$ denotes the total size of the dataset.

\subsection*{LyricBank Dataset Construction.}
LyricBank is a continuously pre-trained corpus focused on song lyrics, aimed at enhancing the model’s understanding of lyrics. To build this dataset, we collected lyrics information from approximately 2.5 million songs from NetEase Music.

\textbf{I. Data Scraping.} During the data collection process, we first used NetEase Music’s API to retrieve metadata for the songs, including song name, artist, album information, and lyrics text. Since the lyrics data might have duplication, missing information, or poor quality, we performed strict cleaning and filtering. 

\textbf{II. Data Cleaning.} We first removed duplicate lyrics entries to ensure that each song in the dataset has unique lyrics. Then, we eliminated songs with missing lyrics or very few lyrics, such as instrumental tracks or songs with only humming parts. Additionally, we filtered out lyrics that contained irrelevant characters or formatting errors to ensure the quality of the data. To maintain diversity, we kept lyrics of different styles, themes, and languages. After multiple rounds of filtering, we retained approximately 1.8 million songs from the original 2.5 million.

\textbf{III. Language Diversity.} To help the model understand and generate lyrics in multiple languages, we included lyrics in four languages: English, Chinese, Korean, and Japanese. The distribution of these lyrics in the dataset is approximately 2:2:1:1, ensuring linguistic richness.

\textbf{IV. General Data Supplementation.} In addition to lyrics, we also added other types of general knowledge data to maintain the model's general understanding. This data includes daily conversations, math problems, code snippets, etc., sourced from public datasets like Wikipedia, LinkSoul, and Falcon. We mixed this general data with lyrics data at a 4:1 ratio to ensure the model retains a balanced understanding of various fields while focusing on lyrics.

\subsection*{LyricBank-SFT Dataset Construction.}
LyricBank-SFT is a dataset designed for instruction fine-tuning, containing 775k samples and covering 9 tasks related to lyrics and playlists, aimed at further enhancing the model's expertise in the music domain. These tasks include:

\textbf{1. User-query rewrite for playlist recommendation:} Rewrite the user's query for the playlist into a specific general description of the playlist.

\textbf{2. Playlist to tags:} Generate tags for the playlist based on its title, song list, and meta description.

\textbf{3. Playlist to description:} Generate description for the playlist based on its title, tags, and song list.

\textbf{4. Tell a story for lyrics:} Narrate the story behind the song based on its lyrics, such as the inspiration for its creation, and the story reflected in the song.

\textbf{5. Lyrics continuation:} Continue writing the second half of the lyrics based on the first half.

\textbf{6. Generate lyrics from keywords:} Write lyrics based on given keywords such as genre and characteristics.

\textbf{7. Infer artist from lyrics:} Infer the singer's name based on the given lyrics.

\textbf{8. Generate song title for lyrics:} Create a song title based on the given lyrics.

\textbf{9. Emotion detection:} Capture the emotion of the song based on the lyrics.

The dataset was built by collecting playlists and lyrics data from NetEase Music and other platforms, followed by task-specific labeling and organization. After filtering for quality, we selected 775k high-quality samples for the LyricBank-SFT dataset.

\subsection*{Preference Dataset Construction.}
Preference Set is a dataset designed to optimize model performance. Due to challenges and ethical concerns in collecting real user preference data, we adopted an alternative approach. We used two models with different performance levels to generate samples. The output from the higher-performing model (e.g., Deepseek model) is labeled as "preferred," while the output from the lower-performing model (e.g., Yi-1.5-6B-Chat model) is labeled as "non-preferred." This generated 1k samples per task, forming the Preference Set dataset. During the model training process, we used Direct Preference Optimization (DPO) to fine-tune the model and improve its performance across tasks.

\subsection*{Dataset Validation}

For large datasets, manual labeling is not practical. To address this, we employed random sampling to validate the dataset, selecting representative samples for manual verification. These samples were chosen to capture the overall characteristics of the dataset, and we conducted multiple rounds of validation to ensure its quality. Furthermore, the quality of the dataset directly influences the effectiveness of model training. Experimental results demonstrated that SongSage outperforms other LLMs on several tasks, which serves as an indirect validation of the dataset's quality.

\subsection*{Analysis of Potential Model Bias in SongSage}

The development of SongSage, while achieving significant advancements in textual music AI, necessitates a thorough examination of potential biases that may arise from the model's training data and methodology. Given that SongSage is primarily trained on lyrics and playlist descriptions, biases may emerge in several dimensions:

I. Cultural and Genre Bias: The dataset used for pretraining may not be equally representative of diverse musical cultures and genres. This could result in the model performing better on mainstream or overrepresented genres while underperforming on niche or underrepresented musical styles.

II. Lyrical Content Bias: Lyrics often reflect specific themes, emotions, or social contexts. If the training data is not sufficiently balanced, the model may exhibit biases towards certain emotional tones or thematic content, potentially leading to skewed outputs in tasks such as emotion detection or lyric generation.

III. User Query Bias: The PlaylistSense dataset, which evaluates the model's ability to understand user queries, is generated based on predefined perspectives. This may inadvertently introduce biases related to the types of queries or user intents that the model is exposed to during training.

To address these potential biases, we plan to conduct a detailed analysis of the training data to identify any imbalances or underrepresented categories in the future work. Additionally, we will implement balanced data augmentation strategies to ensure that the model is exposed to a diverse range of musical genres, cultures, and lyrical themes. This will help mitigate biases and improve the model's fairness and generalizability across different contexts.

\subsection*{Future Work \& Multimodal Integration}

While the current version of SongSage excels in text-based music tasks, we acknowledge the necessity to expand its capabilities to address more comprehensive music AI tasks. Music, by its nature, is multimodal, encompassing audio, text, and symbolic notations. Future work will focus on integrating these additional modalities into SongSage's architecture, thereby enhancing its overall understanding of music.

Specifically, we plan to incorporate audio features (e.g., melodies, harmonies) and symbolic notations (e.g., musical scores) into the model's training pipeline. This multimodal approach will enable SongSage to perform tasks that require a holistic understanding of music, such as music recommendation based on audio features, generation of music videos, and composition of new pieces that align with both lyrical and melodic contexts. By leveraging recent advancements in multimodal learning, we aim to develop a more powerful and versatile music AI model that can better serve the diverse needs of the music community.

In summary, addressing potential biases and expanding SongSage's capabilities through multimodal integration are key priorities for future work. We are committed to developing a more equitable and comprehensive music AI model that can effectively handle a wide range of music-related tasks.

\clearpage
\begin{table}[!h]
\centering
\begin{tabular}{ll}
\hline
\textbf{Hyperparameter}        & \textbf{Setting}  \\ 
\hline  
Batch Size                     & 8                     \\ 
Number of GPUs                 & 16 (2×8)             \\
Epoch                          & 1                     \\
Precision                      & BF16                  \\
Max Context Length             & 2048                  \\ 
Weight Decay                   & 0.1                   \\
Warmup Ratio                   & 0.03                  \\ 
Learning Rate                  & 3e-5                  \\
Gradient Clipping Range        & 1.0                   \\ 
Gradient Accumulation Steps    & 8                     \\
Optimizer                      & AdamW                 \\
LR Scheduler                   & Cosine               \\
Total Steps               & 7147 \\
\hline
\end{tabular}
\caption{Hyperparameters during Pretraining.}
\label{tab:pretrain_hyperparameters}
\end{table}

\begin{figure}[!h]
    \centering
    \includegraphics[width=0.7\linewidth]{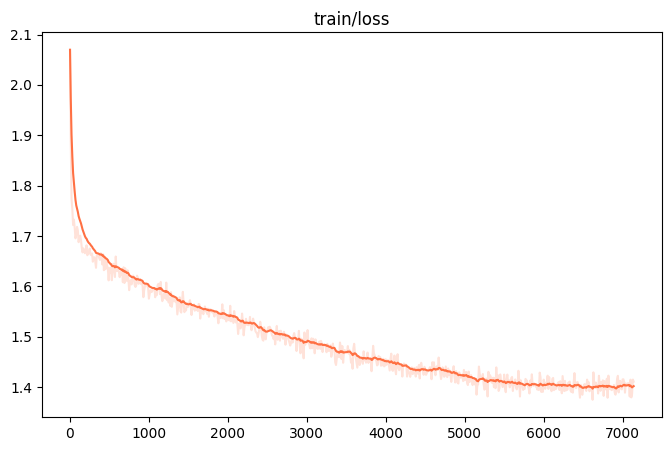}
    \caption{Pretrain: Training Loss Curve.}
    \label{fig:enter}
\end{figure}

\begin{figure}[!h]
    \centering
    \includegraphics[width=0.7\linewidth]{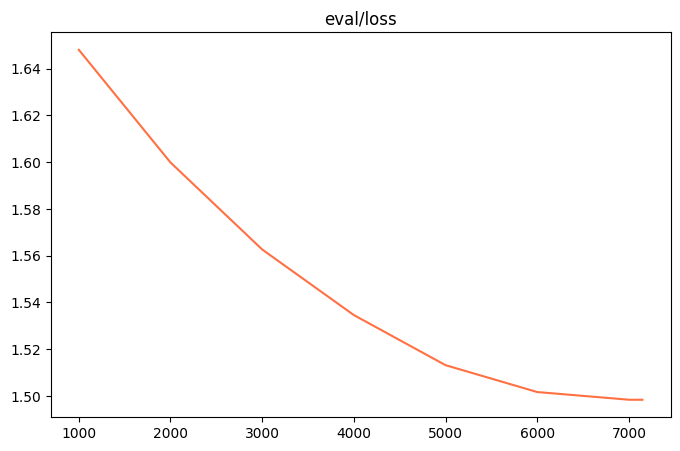}
    \caption{Pretrain: Eval Loss Curve.}
    \label{fig:enter-1}
\end{figure}

\newpage
\begin{table}[!h]
\centering
\begin{tabular}{ll}
\hline
\textbf{Hyperparameter}        & \textbf{Setting}  \\ 
\hline  
Batch Size                     & 8                     \\ 
Number of GPUs                 & 16 (2×8)             \\
Epoch                          & 1                     \\
Precision                      & BF16                  \\
Max Context Length             & 2048                  \\ 
Weight Decay                   & 0.1                   \\
Warmup Ratio                   & 0.03                  \\ 
Learning Rate                  & 1e-4                  \\
Gradient Clipping Range        & 1.0                   \\ 
Gradient Accumulation Steps    & 8                     \\
LoRA Rank                      & 32                    \\ 
LoRA Alpha                     & 64                    \\ 
LoRA Dropout                   & 0.05                  \\
LoRA Target Modules           & ALL                   \\
LR Scheduler                 & Cosine                \\
\hline
\end{tabular}
\caption{Hyperparameters during Instruction Tuning.}
\label{tab:instruction_tuning_hyperparameters}
\end{table}

\begin{figure}[!h]
    \centering
    \includegraphics[width=0.7\linewidth]{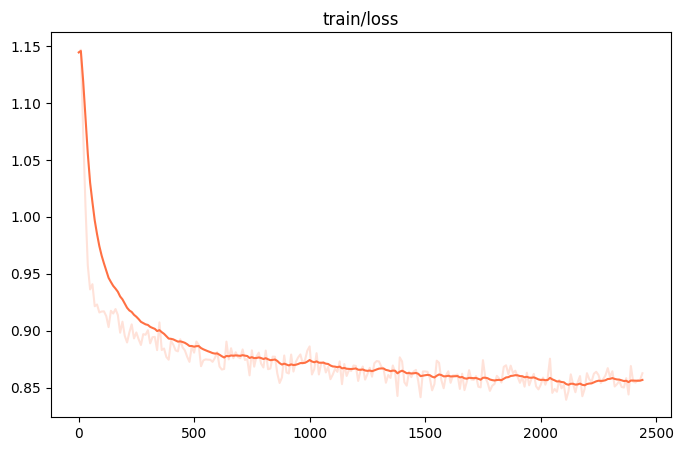}
    \caption{SFT: Training Loss Curve.}
    \label{fig:enter-2}
\end{figure}

\begin{figure}[!h]
    \centering
    \includegraphics[width=0.7\linewidth]{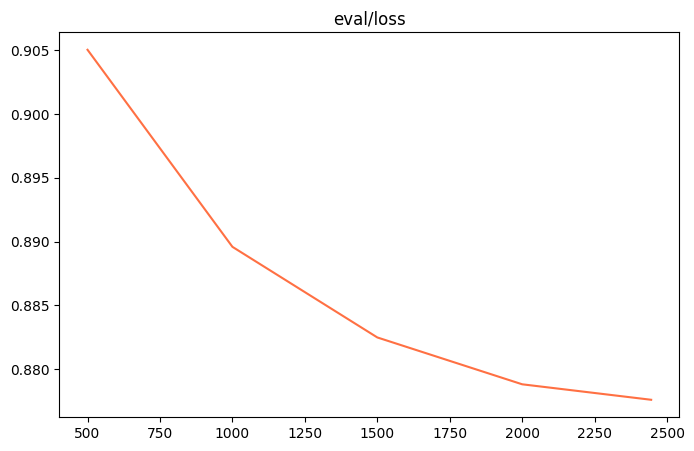}
    \caption{SFT: Eval Loss Curve.}
    \label{fig:enter-3}
\end{figure}

\newpage
\begin{figure*}[!h]
    \centering
    \includegraphics[width=0.7\linewidth]{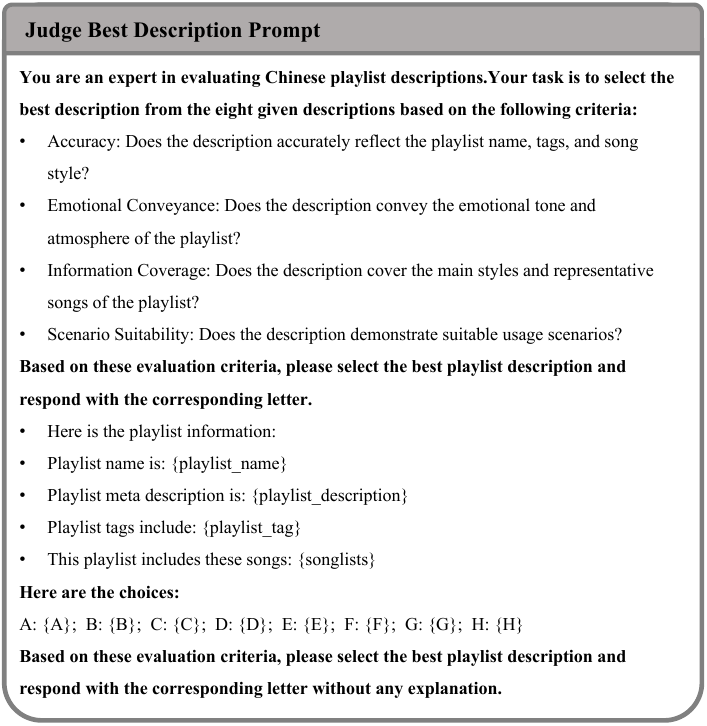}
    \caption{Best Description Judge Prompt.}
    \label{fig:enter-label}
\end{figure*}
\begin{figure*}[!h]
    \centering
    \includegraphics[width=0.7\linewidth]{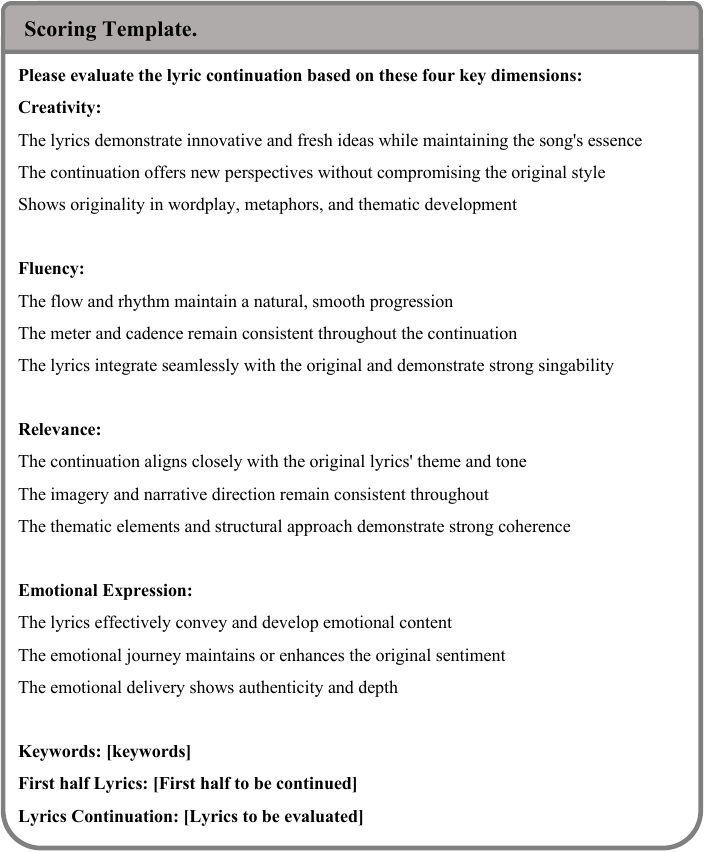}
    \caption{Lyrics Evaluation Prompt.}
    \label{fig:enter-label}
\end{figure*}

\begin{table*}[!t]
\centering
\renewcommand{\arraystretch}{1.1}  
\small
\resizebox{0.88\textwidth}{!}{
\begin{tabular}{l|cc|cc|cc|cc|cc}
\toprule
& \multicolumn{10}{c}{\textbf{Music-Based Perspective}} \\
\cmidrule(lr){2-11}
\textbf{Model} & \multicolumn{2}{c|}{Culture} & \multicolumn{2}{c|}{Emotion} & \multicolumn{2}{c|}{Era} & \multicolumn{2}{c|}{Scene} & \multicolumn{2}{c}{Type} \\
\cmidrule(lr){2-3} \cmidrule(lr){4-5} \cmidrule(lr){6-7} \cmidrule(lr){8-9} \cmidrule(lr){10-11}
& {HR@1} & {HR@5} & {HR@1} & {HR@5} & {HR@1} & {HR@5} & {HR@1} & {HR@5} & {HR@1} & {HR@5} \\
\midrule
\rowcolor{gray!10} Llama-3-8B-Instruct & 1.62 & 4.18 & 1.58 & 4.20 & 1.80 & 3.80 & 1.56 & 3.66 & 2.84 & 6.60 \\
InternLM2.5-7B-Chat & 0.02 & 0.12 & 0.04 & 0.12 & 0.04 & 0.14 & 0.02 & 0.10 & 0.02 & 0.20 \\
\rowcolor{gray!10} Qwen2.5-7B-Instruct & \underline{3.48} & 8.48 & 3.08 & 7.32 & \underline{2.90} & 6.74 & 2.70 & 6.64 & 6.20 & 13.54 \\
GLM4-9B-Chat & 2.82 & 6.40 & 2.90 & 6.48 & 2.38 & 5.50 & 3.18 & 6.60 & 6.02 & 12.98 \\
\rowcolor{gray!10} Claude-3.5-Sonnet& \textbf{3.70} & \textbf{9.94} & \textbf{4.76} & \textbf{12.84} & \textbf{3.66} & \textbf{9.86} & \textbf{5.32} & \textbf{13.60} & \textbf{10.76} & \textbf{25.90} \\
GPT-4o & 2.93 & \underline{9.70} & \underline{2.93} & \underline{8.07} & 2.13 & \underline{7.67} & \underline{3.93} & \underline{11.13} & \underline{7.13} & \underline{16.13} \\
\midrule[\heavyrulewidth]
& \multicolumn{10}{c}{\textbf{User-Based Perspective}} \\
\cmidrule(lr){2-11}
\textbf{Model} & \multicolumn{2}{c|}{Characteristic} & \multicolumn{2}{c|}{Emotion} & \multicolumn{2}{c|}{Habit} & \multicolumn{2}{c|}{Social Role} & \multicolumn{2}{c}{Psychology} \\
\cmidrule(lr){2-3} \cmidrule(lr){4-5} \cmidrule(lr){6-7} \cmidrule(lr){8-9} \cmidrule(lr){10-11}
& {HR@1} & {HR@5} & {HR@1} & {HR@5} & {HR@1} & {HR@5} & {HR@1} & {HR@5} & {HR@1} & {HR@5} \\
\midrule
\rowcolor{gray!10} Llama-3-8B-Instruct & 1.78 & 3.90 & 0.72 & 2.06 & 1.32 & 2.88 & 2.50 & 5.44 & 0.44 & 1.32 \\
InternLM2.5-7B-Chat & 0.02 & 0.08 & 0.02 & 0.08 & 0.00 & 0.06 & 0.06 & 0.14 & 0.00 & 0.12 \\
\rowcolor{gray!10} Qwen2.5-7B-Instruct & 2.94 & 7.18 & 1.18 & 3.60 & \underline{2.24} & 5.32 & \underline{5.08} & 10.84 & 1.06 & 3.32 \\
GLM4-9B-Chat & \underline{3.12} & 6.10 & \underline{1.26} & 3.26 & 2.06 & 4.80 & 4.76 & 10.00 & 1.36 & 3.70  \\
\rowcolor{gray!10} Claude-3.5-Sonnet & \textbf{3.16} & \textbf{9.94} & \textbf{1.74} & \textbf{12.84} & \textbf{2.54} & \textbf{9.86} & \textbf{6.28} & \textbf{13.60} & \underline{1.54}& \underline{4.34} \\
GPT-4o & 2.33 & \underline{7.47} & 0.93 & \underline{4.47} & 1.67 & \underline{6.00} & 4.13 & \underline{12.07} & \textbf{2.33} & \textbf{6.00} \\
\bottomrule
\end{tabular}
}
\caption{HR@1(\%) and HR@5(\%) of GPT-4o, Claude 3.5 Sonnet, and advanced open-source models across 10 perspectives on the PlaylistSense dataset in English. Bold denotes the best performance, and underlined results indicate runner-ups.}
\label{benchmark_results_en}
\end{table*}

\newpage
\begin{table*}[!t]
\centering
\resizebox{0.8\textwidth}{!}{
\small
\begin{tabular}{llrrlc}
\toprule
Datasets & Sourced from & Tokens & \#Samples & Category & Format \\
\hline
pile & public dataset & 0.83B & 18K & general & article \\
Falcon-RefinedWeb & public dataset & 0.80B & 101K & general & article \\
Wikipedia & public dataset & 0.39B & 588K & general & article \\
OpenChat & public dataset & 62.44M & 43K & general & chat \\
LinkSoul & public dataset & 0.6B & 1.5M & general & chat \\
GPT4-Alpaca & public dataset & 9.77M & 49K & general & chat \\
Dolly & public dataset & 3.12M & 14K & general & chat \\
IrishMAN & public dataset & 0.23B & 868K & music score & chat \\
KernScores & public dataset & 2.76M & 10K & music score & chat \\
JSB Chorales & public dataset & 0.44M & 349 & music score & chat \\
synthetic music chat* & public dataset & 0.54B & 50K & music score & chat \\
music knowledge** & Generated & 0.22B & 255K & music verbal & chat \\
music summary** & Generated  & 0.21B & 500K & music verbal & chat \\
GSM8k & public dataset & 1.68M & 7K & math & chat \\
math & public dataset & 7.03M & 37K & math & chat \\
MathInstruct & public dataset & 55.50M & 188K & math & chat \\
Camel-Math & public dataset & 27.76M & 50K & math & chat \\
arxiv-math-instruct-50k & public dataset & 9.06M & 50K & math & chat \\
Camel-Code & public dataset & 0.13B & 366K & code & chat \\
OpenCoder & public dataset & 36.99M & 28K & code & chat \\
Lyrics & NetEase Music & 1.03B & 1.8M & Lyrics & Sequential \\
\bottomrule
\end{tabular}
}
\caption{Distributions of LyricBank.}
\end{table*}

\end{document}